\pdfoutput=1

\documentclass[11pt]{article}

\usepackage{acl}

\usepackage{times}
\usepackage{latexsym}
\usepackage{CJKutf8}

\usepackage[T1]{fontenc}

\usepackage[utf8]{inputenc}

\usepackage{microtype}

\usepackage{inconsolata}

\usepackage{booktabs} 
\usepackage{bm}
\usepackage{multirow}
\usepackage{xspace}
\usepackage[tight,footnotesize]{subfigure}
\usepackage{algorithm}  
\usepackage{amsmath}
\usepackage{enumitem}
\usepackage{algorithmic}
\usepackage{balance}
\usepackage{cancel}
\usepackage{graphicx}
\usepackage{enumerate}
\usepackage{amssymb}
\usepackage{bbding}
\usepackage{float}
\usepackage{verbatimbox}

\usepackage{colortbl}

\newcommand{\ie}{\emph{i.e.,}\xspace}

\newcommand{\eg}{\emph{e.g.,}\xspace}

\newcommand{\wrt}{\emph{w.r.t.}\xspace}

\newcommand{\model}{\textsc{MetRag}}
\newcommand{\comb}{\textsc{MetRag w/o Comb}}
\newcommand{\as}{\textsc{MetRag w/o AS}}

\title{Similarity is Not All You Need: Endowing Retrieval-Augmented Generation with Multi–layered Thoughts}

\author{Chunjing Gan\quad Dan Yang\quad Binbin Hu\quad Hanxiao Zhang \quad Siyuan Li\quad  Ziqi Liu \\ 
\bf Yue Shen\quad Lin Ju\quad Zhiqiang Zhang\quad Jinjie Gu\quad Lei Liang\quad Jun Zhou\footnotemark[2] \\
        Ant Group \\
        jun.zhoujun@antgroup.com
        }

\begin{document}
\maketitle

\begin{abstract}
In recent years, large language models (LLMs) have made remarkable achievements in various domains. However, the untimeliness and cost of knowledge updates coupled with hallucination issues of LLMs have curtailed their applications in knowledge-intensive tasks, where retrieval-augmented generation (RAG) can be of help. Nevertheless, existing retrieval-augmented models typically use similarity as a bridge between queries and documents and follow a retrieve-then-read procedure.
In this work, we argue that similarity is not always the ``panacea'' and totally relying on similarity would sometimes degrade the performance of retrieval-augmented generation.
To this end, we propose \textbf{{\model}}, a \textbf{M}ulti–lay\textbf{E}red \textbf{T}houghts enhanced \textbf{R}etrieval-\textbf{A}ugmented \textbf{G}eneration framework.
To begin with, beyond existing similarity-oriented thought, we embrace a small-scale utility model that draws supervision from an LLM for utility-oriented thought and further come up with a ``smarter'' model by comprehensively combining the similarity- and utility-oriented thoughts.
Furthermore, given the fact that the retrieved document set tends to be huge and using them in isolation makes it difficult to capture the commonalities and characteristics among them, we propose to make an LLM as a task-adaptive summarizer to endow retrieval-augmented generation with compactness-oriented thought.
Finally, with multi-layered thoughts from the precedent stages, an LLM is called for knowledge-augmented generation.
Extensive experiments on knowledge-intensive tasks have demonstrated the superiority of {\model}.
\end{abstract}

\section{Introduction}
In recent years, large language models (LLMs) such as ChatGPT \cite{chatgpt2023}, GPT4 \cite{gpt42023}, Llama \cite{touvron2023llama2} have made remarkable achievements in a variety of tasks due to their marvelous capability in language comprehension and generation \cite{contextfaith2023zhou}.
However, the untimeliness and cost of knowledge updates \cite{realtimeqa2022} together with hallucinations issues of LLMs~\cite{hallucinationsurvey2023} have curtailed their applications in knowledge-intensive tasks to a large extent \cite{whennotto2023mallen}, where retrieval-augmented generation (RAG) approaches that prepending documents to the query without updating the underlying language models would come in handy \cite{iclretrieve2023,ragtutorial2023asai}.
Nevertheless, existing retrieval-augmented generation approaches are typically similarity-based \cite{flare2023}, \ie they retrieve documents from external corpus based on similarity. Then, the retrieved documents are prepended as context for LLMs at once or independently for generation augmentation \cite{replug2023}. In total, these approaches have been found to outperform purely parametric
LLMs \cite{flare2023}, especially in some knowledge-intensive generation tasks \cite{nq2019}.

\begin{figure}[h]
\centering
\includegraphics[width=1.0\columnwidth]{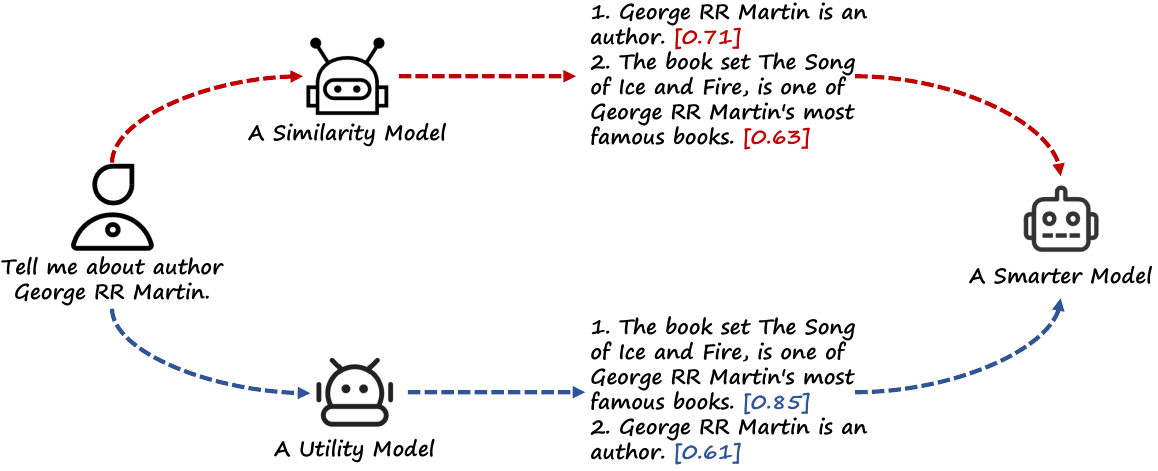}
\caption[Caption for examples]{A toy example illustrating the difference between similarity and utility, where the score of similarity model is given by BGE\protect\footnotemark.  Can we reunite the virtues of both worlds and come up with a better model?}
\label{fig:eg}
\end{figure}

In this work, we argue that similarity is not always the ``panacea'' for retrieval-augmented generation and totally relying on similarity would sometimes degrade the performance.
As is shown in the upper part of Figure \ref{fig:eg}, when a user types in the query ``Tell me about author George RR Martin'', a similarity-driven retrieval system would rank the documents in a given corpus according to similarity metrics, \ie the semantic relevance or TF-IDF based metric \cite{bm252009}. Even though the retrieved documents barely provide useful information, \eg ``George RR Martin is an author'', it would rank higher due to high similarity score and the document that states the publications ``The Song of Ice and Fire'' of George RR Martin with higher information gain would rank lower due to inadequate low similarity score. Besides, given the fact that the retrieved documents are often more than one, using them in isolation due to the context limitation of LLMs \cite{replug2023} or simply aggregating the Top-$k$ document without considering the relationships between them makes it difficult to capture the commonalities and characteristics among them and even confuse LLMs due to excessive text length thus incurring information loss and probably performance degradation \cite{whennotto2023mallen}.

\footnotetext{https://github.com/FlagOpen/FlagEmbedding}

Given the above limitations in current approaches, beyond similarity we aim to endow retrieval-augmented generation with multi-layered thoughts (\ie utility- and compactness-oriented thoughts) for performance boosting.
However, the solution is quite non-trivial, which needs to tackle the following essential challenges:
(\textbf{C1}) To train a model that is capable of perceiving utility-oriented thoughts rather than solely similarity, external labeled data is required. 
However, it is hard to obtain external labeled data for guiding the learning process with explicit supervision. Though LLMs can serve as data annotators and come up with high-quality corpus for model training and have demonstrated their tremendous capability in many circumstances, the innate uncontrollable and instability characteristics would sometimes deteriorate the performance.
(\textbf{C2}) With high-quality retrieved documents, to reduce the burden that dozens of documents impose on LLMs and better capture the commonalities and characteristics between retrieved documents, document summarization is plausible. However, simple summarization cannot guarantee that the most important information \wrt the input query can be retained, hence there is need to train a summarization model that aligns with the task itself thus possessing compactness-oriented thoughts.

To this end, we propose {\model}. 
In particular, with an LLM serves as supervision on document utility \wrt the input query, we come up with a utility model that aligns itself with the LLM's feedback such that beyond similarity it is also equipped with utility-oriented thoughts.
Considering that an LLM is strong most of time but sometimes would go out of control, yet dense retrievers trained on labelled corpus remain stable with relevance guarantee, though sometimes useless, we further combine similarity- and utility-oriented thoughts and takes the output of a similarity model and a utility model into consideration (\textbf{C1}), as depicted in Figure \ref{fig:eg}. 
Furthermore, to endow the summarization model with compactness-oriented thoughts, we first distill the summary ability from a strong teacher model (\eg GPT4). Afterwards with multiple generated summaries, a subsequent reward model is utilized to further constrain the summarization model to align with the end task (\textbf{C2}). 
With the multi-layered thoughts derived from the precedent stages, an LLM is called for knowledge-augmented generation.
At last, we evaluate the proposed {\model} on multiple knowledge-extensive tasks, extensive experiments and analysis have demonstrated the superiority of the proposed method.
\section{Related work}
\subsection{Retrieval-Augmented Generation}
Retrieval-Augmented Generation (RAG) approaches \cite{ragsurvey2023,ragtutorial2023asai,iclretrieve2023} tend to enhance LLMs following a retrieve-then-read pipeline when given the input query, which first retrieve a set of documents from the external corpus, and then utilize the retrieved documents as side information for an LLM to make the final prediction. This pipeline has demonstrated their superior performance, especially in knowledge-intensive tasks by fine-tuning or directly prepending to LLMs \cite{promptcap2022,atlas2022}.
However, the essence of RAG is the quality of retrieved passages, yet plenty of existing approaches rely on similarity-based retrieval \cite{recomp2023,whennotto2023mallen} that tend to ignore the underlying utility of associated passages. Though Self-RAG, REPLUG \cite{selfrag2023asai,replug2023} have made progress in introducing the power of LLMs in augmenting retrieval ability and even achieve adaptive retrieval, purely depending on the capability of LLMs can  
be dangerous since the innate ``uncontrollable'' characteristics of LLMs would sometimes degrade the performance. Furthermore, they tend to ignore the potential inner relationships among retrieved passages and utilize them in isolation, which deteriorate the performance.

\subsection{Task-Oriented Summarization}
Large language models (LLMs) \cite{LLMSurvey2023,chatgpt2023,gpt42023} have made remarkable achievements in a variety of domains such as question answering, and summarization. However, calling large-scale commercial LLMs are of high cost and may induce data leakage issue, hence many approaches have been devoted to distilling the ability of large-scale LLMs (\eg ChatGPT \cite{chatgpt2023}) to small-scale LLMs (\eg Llama2 \cite{touvron2023llama2}) and enhance their capabilities for downstream tasks, \eg generate high-quality summary through distillation \cite{summary2023}. However, for knowledge-intensive tasks, simple summarization is far from optimized because it cannot ensure that the most important information relevant to the input query is retained, hence it is necessary to generate the summary associated with downstream tasks. Though RECOMP \cite{recomp2023} has made its step to train the summarization model for enhancing performance in tasks such as question answer and language modeling, it purely designs intricate samples and performs distillation without further aligning strategy for performance boosting, which downgrades its performance in applications.
\section{The proposed approach}\label{sec:model}

\begin{figure*}[h]
    \centering
    \includegraphics[width=1\textwidth]{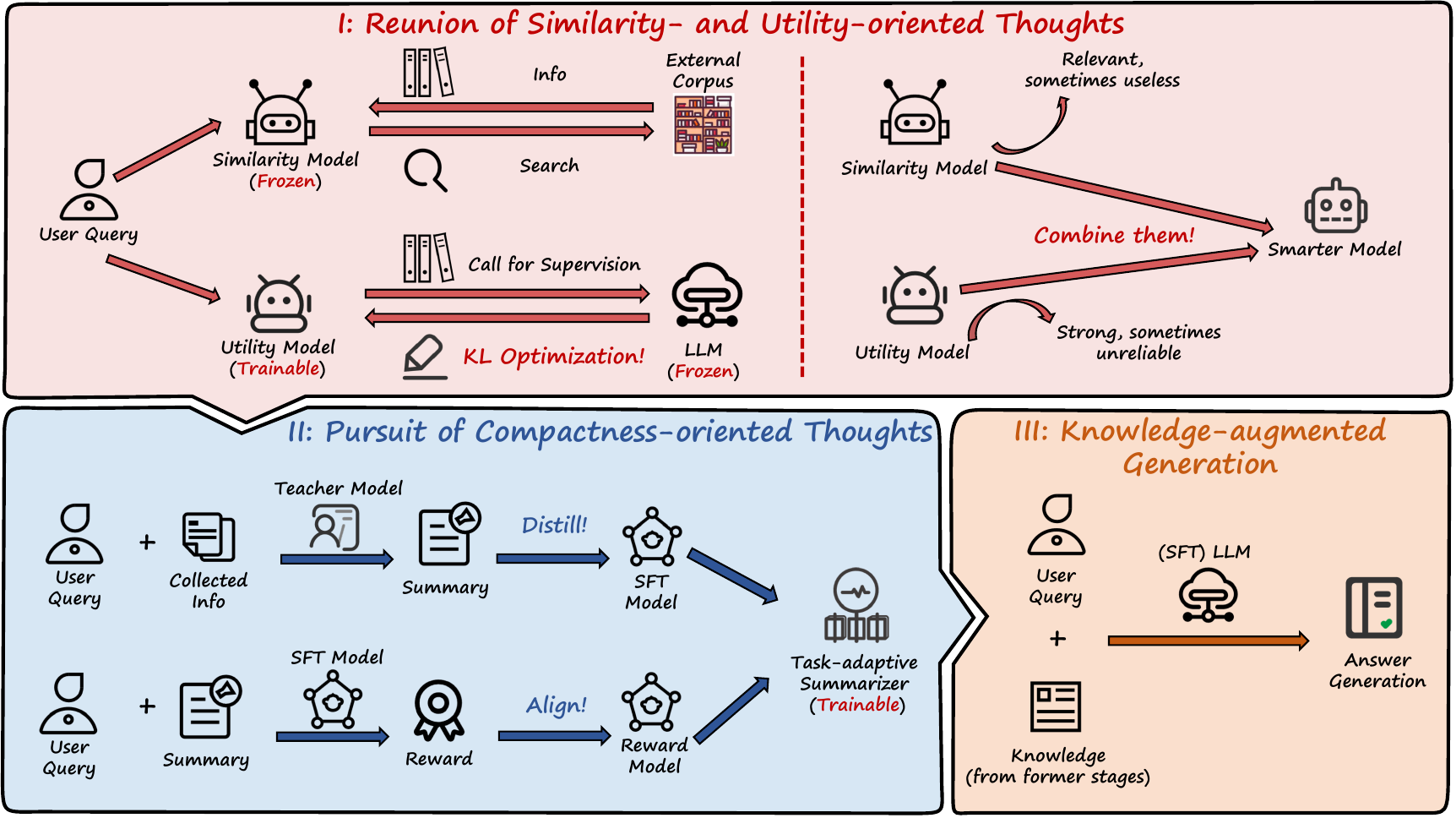}
    \caption{The proposed {\model} framework, where we endow retrieval-augmented generation with multi-layered thoughts from Stage I and II, and utilize the derived knowledge in Stage III for answer generation.}
    \label{fig:framework}
\end{figure*}

\subsection{Overview}
In this work, we propose {\model}, as shown in Figure \ref{fig:framework}. {\model} endows retrieval-augmented generation with multi-layered thoughts by firstly embracing LLM's supervision for utility-oriented thoughts and combining similarity and utility of documents for performance boosting (\textbf{I}, detailed in Section \ref{sec:combination}) and further pursuing compactness-oriented thoughts via a task-adaptive summarizer (\textbf{II}, detailed in Section \ref{sec:summary}), finally incorporating the derived multi-layered thoughts for answer generation (\textbf{III}, detailed in Section \ref{sec:task}). 

\subsection{A Tale of Two ``Models''}\label{sec:combination}

\subsubsection{Similarity Model as an Off-the-shelf Retriever}\label{sec:model_similarity}
Given an input query $q$, an off-the-shelf dense retriever $\mathcal{R}$ is incorporated to map the input to a low-dimensional embedding $\mathbf{E}(\cdot)$ such that we can efficiently retrieve relevant documents \wrt query $q$ from a given corpus $\mathcal{D} = \{d_1, d_2, ..., d_n\}$ via a predefined similarity metric $\mathcal{M}$ (which could be cosine similarity and so on), where the similarity score between the query and document can be computed as follows:
\begin{equation}
    s(q,d) = \mathcal{M}(\mathbf{E}(q),\mathbf{E}(d))
\end{equation}
With the derived similarity scores, the Top-$k$ documents $\mathcal{D^*}$ that have the highest similarity scores \wrt input query $q$ are retrieved for enhancing follow-up tasks.

\subsubsection{LLM's Supervision Empowered Utility Model}\label{sec:model_utility}
In this work, we argue that similarity is not always the ``panacea'' for information retrieval and totally relying on similarity would sometimes degrade the performance of retrieval-augmented generation approaches. As is illustrated in Figure \ref{fig:eg}, document with higher information gain to the input query would rank lower due to inadequate low similarity score.  Inspired by the great success of LLMs in a variety of tasks, we aim to incorporate an LLM for supervision on document utility (In this paper, we define the utility of one document \wrt a question by its usefulness in assisting an LLM to answer this question, which is modelled by the normalization of the probability of generating correct answers with a specific LLM.) \wrt the input query to endow the retriever $\mathcal{R}$ with utility-oriented thoughts.

To start with, given the Top-$n$ documents set $\mathcal{D^\bullet}$ ($\mathcal{D}^\bullet \subseteq  \mathcal{D^*}$) that a retriever $\mathcal{R}$ considers most similar to the input query $q$, the approximated similarity likelihood of each document $d$ can be formalized as follows:
\begin{equation}
    \mathbf{P}_\mathcal{R}(d|q) = \frac{e^{\mathcal{M}(q,d)/\tau}}{\sum_{d \in \mathcal{D}^\bullet} e^{\mathcal{M}(q,d)/\tau}},
\end{equation}
where $\tau$ is the temperature hyperparameter that controls the smoothness of probability distribution, with a higher $\tau$ produces a softer probability distribution while a lower $\tau$ results in a "harder" max operation and $\mathcal{D}^\bullet$ denotes the documents pool for model training. 
To endow the retriever $\mathcal{R}$ with LLM's insights on retrieval utility, we further incorporate an LLM as the supervision signal on document utility of $d$ \wrt the input query $q$ such that beyond similarity the trained utility model can take the utility a document provides into consideration, which we can formalize as follows: 
\begin{equation}
    \mathbf{P}_\mathcal{U}(d|q,y) = \frac{e^{\mathbf{P}_\textit{LLM}(y|q,d)/\tau}}{\sum_{d \in \mathcal{D}^\bullet} e^{\mathbf{P}_\textit{LLM}(y|q,d)/\tau}}.
\end{equation}

Finally, we break down the barriers of both sides and push the similarity distribution toward LLM-supervision enhanced utility distribution by minimizing the KL-divergence of these two distributions as follows:
\begin{equation}
    \mathcal{L}_\textit{U} = \frac{1}{\left| \mathcal{Q} \right|} \sum_{q \in \mathcal{Q}}\mathbf{KL}({\mathbf{P}_\mathcal{R}(\cdot)}_{d \in {\mathcal{D}^\bullet}}||{\mathbf{P}_\mathcal{U}(\cdot)}_{d \in {\mathcal{D}^\bullet}}),
\end{equation}
where $\mathcal{Q}$ is a set of queries. Due to the reason that the LLM is served as the external supervision, during training the parameter of the LLM is frozen and we only update the parameters of retriever $\mathcal{R}$ and finally come up with the utility model $\mathcal{U}$, thus the overall optimization process remains lightweight. Besides, we further add an empty string $es$ to $\mathcal{D}^\bullet$ in the training process to enable the utility model with the ability to judge whether introducing a document for a query can promote the utility or not thus achieving selective retrieval.

\subsubsection{Reunion of Similarity- and Utility- oriented Thoughts}
Considering that an LLM would sometimes go out of control and produces unreliable supervision signals which degrade the model performance sometimes yet dense retrievers trained on accurately labelled corpus remain stable with relevance guarantee, though sometimes useless, we further combine similarity- and utility-oriented thoughts and take the output of a similarity model into consideration via an integration strategy $\mathcal{I}(\cdot)$. Therefore, the final score between document $d$ and the input query $q$ can be defined as follows:
\begin{equation}
    \begin{split}
        f(q, d) & = \mathcal{I}(\mathcal{R}(q,d), \mathcal{U}(q,d)) \\
        & = \begin{cases}
            1.0, \quad \mathcal{R}(q,d) \geq  \zeta^{\mathcal{R}}(k^{\mathcal{R}}) \\
            1.0, \quad \mathcal{U}(q,d)   \geq  \zeta^{\mathcal{U}}(k^{\mathcal{U}}) \\
            0.0, \quad \text{otherwise}
        \end{cases}
    \end{split}
\end{equation}
where $\zeta^{\mathcal{R}}(k^{\mathcal{R}})$ and $\zeta^{\mathcal{U}}(k^{\mathcal{U}})$ find value of the $k^{\mathcal{R}}$-largest similarity score and $k^{\mathcal{U}}$-largest utility score among retrieved documents through similarity and utility, respectively. Finally, only documents with $f(q, d) = 1.0$ are permitted to proceed to the subsequent stages. With the derived final score, we can easily obtain the current documents set $\mathcal{D}^\Delta$ \wrt input query $q$ for follow-up tasks.

\subsection{Pursuit of Compactness-oriented Thoughts}\label{sec:summary}
Given the fact that the retrieved documents are often more than one, using them in isolation due to the context limitation of LLMs or simply aggregating the Top-$k$ document without considering the relationships between them makes it difficult to capture the commonalities and characteristics among them and even confuse LLMs thus incurring information loss and probably performance degradation, where text summarization can be of help.
However, simple summarization cannot ensure that the most important information relevant to the input query is retained, therefore it is necessary to train a summarization model that aligns with the task itself. Hence, we propose the \textbf{Task-adaptive Summarizer}, which not only reduces the computational costs in end tasks but also relieves the burden of LLMs to identify relevant information in a large chunk of retrieved documents to obtain compactness-oriented thoughts.

\subsubsection{Distilling from Strong Teacher Models}
To initiate the summarization process, we design instructions via randomly sampled queries and retrieved documents. 
To start with, given query $q$, we harness the expertise of a sophisticated teacher model $\mathcal{T}$, \eg GPT4, to extract summarization proficiencies that facilitate the creation of a preliminary summary $\mathbb{S}_q$ and compile an initial corpus $\mathcal{C}_\mathcal{I}$ consisting of Instruction-Summary pairs
$\left\langle \mathbb{I}_q^s,\mathbb{S}_q \right\rangle$, where $\mathbb{I}_q^s$ and $\mathbb{S}_q$ are defined as follows:
\begin{equation}
\mathbb{I}_q^s= \textsc{Template}(\mathbb{I}^s,q,[d_q^1,d_q^2,\cdots,d_q^n]),
\end{equation}
\begin{equation}
\mathbb{I}_q^s \xrightarrow{\mathcal{T}} \mathbb{S}_q,
\end{equation}
where $\mathbb{I}^s$ is the summarization instruction, $\textsc{Template}$ is the summarization prompt template (detailed in Section \ref{app:prompt}) and $d_q^i$ is the $i$-th document retrieved for query $q$ from former stage. 
Next, we apply Lora tuning to meticulously refine an open-source student model, \eg Llama2, which results in an initial summarizer model $\mathcal{\pi}^{\text{SFT}}$ tailored for the end task.

\subsubsection{Alignment via End-task Feedback}
To ensure the faithfulness of the summarizer to the end task, inspired by the principles of the DPO \cite{rafailov2023direct}, we incorporate the LLM’s performance on the end task as reward for the summarizer. 
With model $\mathcal{\pi}^{\text{SFT}}$, given query $q$ we adeptly produce concise summaries of the retrieved documents as $\mathbb{S}_q$ and generate prompt $\mathbb{X}_q$ to obtain end task response $\mathbb{Y}_q$ and the label $\mathbb{Z}_q$ indicating whether the response is correct or not, which accords with the following distribution:
\begin{equation}
p(\mathbb{Z}_q=1|\mathbb{X}_q,\mathbb{Y}_q)=\frac{1}{1+{e}^{-r^*(\mathbb{X}_q,\mathbb{Y}_q)}}.
\end{equation}
Formally, we define the training corpus of this aligning process as $\mathcal{C}_\mathcal{A}$ consisting of triplets $(\mathbb{X}_q,\mathbb{Y}_q,\mathbb{Z}_q)$ for each query $q$. 
With a reward model $r_{\phi}$ parameterized by $\phi$ to estimate $r^*$, the binary classification loss can be defined as follows:
\begin{equation}
    \begin{split}
    \mathcal{L}_{RM}(\phi) =\sum_{(\mathbb{X}_q,\mathbb{Y}_q,\mathbb{Z}_q) \in \mathcal{C}_\mathcal{A}} -\mathbb{Z}_q log \sigma(r_{\phi}(\mathbb{X}_q,\mathbb{Y}_q)) - \\
    (1-\mathbb{Z}_q)log \sigma(1-r_{\phi}(\mathbb{X}_q,\mathbb{Y}_q)),
    \end{split}
\end{equation}
where $\sigma$ is the sigmoid function.
Furthermore, we follow the concept of DPO that eschews the need to explicitly estimate the reward model by solving $r_{\phi}$ as a function of language model policy ${\pi}^{\theta}$, which is formalized as follows:  
\begin{equation}
r_{\phi}(\mathbb{X}_q,\mathbb{Y}_q) = \beta log \frac{{\pi}^{\theta}(\mathbb{Y}_q|\mathbb{X}_q)}{{\pi}^{\text{SFT}}(\mathbb{Y}_q|\mathbb{X}_q)}.
\end{equation}

\subsection{Knowledge-augmented Generation}\label{sec:task}
With the input query $q$ and the external knowledge $\mathcal{K}$ derived from the former stages, we can directly call an LLM (can be fine-tuned in a supervised manner using question answering datasets), where its knowledge-augmented generation of answer $a$ can be formalized as follows:
\begin{equation}
a^* = \underset{a}{\arg\max} \; \mathbf{P}(a|q,\mathcal{K}),
\end{equation}
where $\mathbf{P}(a|q,\mathcal{K})$ is the probability of the answer $a$ given the query $q$ and the external knowledge $\mathcal{K}$, and $\arg\max$ denotes the argument of the maximum, i.e., the answer $a$ for which $\mathbf{P}(a|q, \mathcal{K})$ is maximized.
\section{Experiments}\label{sec:exp}

\subsection{Experimental Setup}\label{sec:exp_setup}
\subsubsection{Tasks and Datasets}
We evaluate our proposed {\model} and multiple baselines on a variety of knowledge-intensive public datasets (including general Open-Domain QA: NQ, TriviaQA-unfiltered\footnote{As the test set is not publicly available, we follow the split setting of \cite{realm2020} and use 11,313 queries for testing.}, HotpotQA and entity-centric QA datasets: PopQA\footnote{We test on the long-tail subset that includes 1399 queries with less than 100 Wikipedia page views.}) and evaluate the performance via metrics EM (following \cite{whennotto2023mallen}, we evaluate the performance based on whether the gold answers are included in the model generations rather than strictly exact matching) and F1.
All experiments are conducted in a zero-shot manner, where we provide instructions and retrieved information about tasks without few-shot demonstrations \cite{zeroshot2022}.

\subsubsection{Baselines}
\noindent{\bf Baselines without retrievals.}
We evaluate strong publicly available pre-trained LLMs,
ChatGLM2$_{\textsc{6b}}$~\cite{du2022glm},
Llama2$_{\textsc{7b},\textsc{13b}}$~\citep{touvron2023llama2}, Baichuan$_{\textsc{7b},\textsc{13b}}$~\citep{baichuan2023baichuan2} and Qwen$_{\textsc{7b},\textsc{14b}}$~\citep{qwen2023qwen}; and models trained and reinforced using private data, ChatGPT\footnote{We use GPT-3.5-turbo-16K in our experiment.}~\citep{chatgpt2023}. 

\noindent{\bf Baselines with retrievals.}  
We evaluate models augmented with retrieval only at test time or during training. The first category includes standard RAG baselines, where an LM (\eg Llama2$_{\textsc{7b},\textsc{13b}}$) generates output given the query prepended with the top retrieved documents. The latter category includes approaches that are trained with retrieved passages, including Self-RAG \cite{selfrag2023asai}, RECOMP \cite{recomp2023}\footnote{We report results reported in the original paper.}.

\subsubsection{Settings}\label{sec:exp_setup_settings}
\paragraph{Training details.}
We split all train, validation and test sets of datasets following \cite{atlas2022,selfrag2023asai}.
Our training data includes randomly sampled instruction-following input-output pairs. We randomly sample 50k pairs from training sets of NQ, TriviaQA and HotpotQA to train our model in different stages with Llama2$_{\textsc{13b}}$ as base LLM. 
All experiments are conducted using 4 NVIDIA A100 GPUs.
We train the utility model for 5 epochs with a learning rate of 1e-5, a batch size of 16 for each device, a warm-up ratio of 0.2, the passage window size of 50 and the temperature parameter $\tau$ set to 0.05.
For task-adaptive summarizer and generation model, we utilize open-source Llama Factory \footnote{https://github.com/hiyouga/LLaMA-Factory} and adopt Lora tuning for 1 epoch with a learning rate of 5e-5, a batch size of 4 and a cosine learning rate scheduler.
\paragraph{Retrieval details.}
For NQ, TriviaQA and HotpotQA, we use 2018 English Wikipedia as the external retrieval source while for PopQA, since the 2018 Wikipedia sometimes lacks articles about entities that have been added to Wikipedia recently, we use December 2020 preprocessed Wikipedia corpus as the external retrieval source. We preprocess the external corpus following \cite{atlas2022} \footnote{https://github.com/facebookresearch/atlas}.
As for retriever $\mathcal{R}$, we incorporate BGE owing to its superior performance in a variety of benchmark leaderboards and retrieve up to 5 documents for each input query for testing, where $\mathcal{M}(\cdot)$ is defined as cosine similarity. 

\subsection{Results and Analysis}
\begin{table*}[t]\centering
\setlength{\tabcolsep}{3.5mm}{

\caption{Overall performance evaluation across 4 public datasets. The best results are highlighted in boldface.
} 
\label{tab:overall_result1}
\begin{tabular}{c|c|c|c|c|c|c|c|c}

\toprule
\multirow{2}{*}{Methods}
& \multicolumn{2}{c|}{NQ} 
& \multicolumn{2}{c|}{TriviaQA} 
& \multicolumn{2}{c|}{HotpotQA} 
& \multicolumn{2}{c}{PopQA} 
\\    
\cmidrule{2-9}
 {} 
 & {EM} & {F1} 
 & {EM} & {F1} 
 & {EM} & {F1} 
 & {EM} & {F1}\\ 
\midrule 
\multicolumn{9}{c}{Baselines without retrieval}  
\\
{ChatGLM2$_{\textsc{6B}}$} 
& {11.1} & {8.80} 
& {22.5} & {20.0} 
& {13.5} & {14.6}  
& {18.8} & {14.0}  \\
{Llama2$_{\textsc{7B}}$} 
& {21.5} & {12.3} 
& {43.0} & {29.0} 
& {17.0} & {10.8}  
& {19.6} & {3.36}  \\
{Baichuan2$_{\textsc{7B}}$} 
& {18.0} & {14.0} 
& {39.3} & {35.0} 
& {18.8} & {15.9}  
& {25.1} & {17.8}  \\
{Qwen$_{\textsc{7B}}$} 
& {20.7} & {17.9} 
& {47.3} & {42.1} 
& {21.7} & {19.6}  
& {21.6} & {15.4}  \\
{Llama2$_{\textsc{13B}}$} 
& {32.4} & {8.60} 
& {58.9} & {14.6} 
& {24.1} & {8.53}  
& {19.0} & {2.91}  \\
{Baichuan2$_{\textsc{13B}}$} 
& {20.8} & {20.6} 
& {46.2} & {45.4} 
& {17.7} & {20.5}  
& {23.8} & {16.4}  \\
{Qwen$_{\textsc{14B}}$} 
& {25.9} & {26.9} 
& {53.4} & {53.7} 
& {24.1} & {25.6}  
& {19.7} & {18.7}  \\
{ChatGPT} 
& {37.8} & {42.9} 
& {71.4} & {72.3} 
& {27.9} & {34.2}  
& {25.2} & {25.7}  \\
\multicolumn{9}{c}{Baselines with retrieval} \\
{ChatGLM2$_{\textsc{6B}}$} 
& {42.8} & {28.2} 
& {59.1} & {47.2} 
& {30.3} & {27.1}  
& {60.3} & {43.7}  \\
{Llama2$_{\textsc{7B}}$} 
& {42.2} & {40.6} 
& {65.4} & {62.5} 
& {30.3} & {33.0}  
& {60.3} & {57.8}  \\
{Baichuan2$_{\textsc{7B}}$} 
& {43.4} & {34.3} 
& {65.0} & {54.7} 
& {33.5} & {29.1}  
& {62.9} & {47.0}  \\
{Qwen$_{\textsc{7B}}$} 
& {43.7} & {31.2} 
& {62.1} & {48.3} 
& {30.9} & {24.8}  
& {40.4} & {27.3}  \\
{Llama2$_{\textsc{13B}}$} 
& {\textbf{50.3}} & {21.2} 
& {70.2} & {25.7} 
& {38.2} & {17.3}  
& {62.9} & {15.1}  \\
{Baichuan2$_{\textsc{13B}}$} 
& {44.4} & {41.7} 
& {65.1} & {62.6} 
& {30.2} & {32.4}  
& {61.4} & {53.1}  \\
{Qwen$_{\textsc{14B}}$} 
& {\textbf{50.3}} & {47.5} 
& {69.4} & {67.1} 
& {39.2} & {39.9}  
& {\textbf{64.1}} & {57.6}  \\
{SELF-RAG*$_{\textsc{7B}}$} 
& {-} & {-} 
& {66.4} & {-} 
& {-} & {-}  
& {54.9} & {-}  \\
{SELF-RAG*$_{\textsc{13B}}$} 
& {-} & {-} 
& {69.3} & {-} 
& {-} & {-}  
& {55.8} & {-}  \\
{RECOMP*$_{\textsc{EXT}}$} 
& {36.6} & {44.2} 
& {59.0} & {65.3} 
& {30.4} & {40.1}  
& {-} & {-}  \\
{RECOMP*$_{\textsc{ABS}}$} 
& {37.0} & {45.5} 
& {58.7} & {66.3} 
& {28.2} & {37.9}   
& {-} & {-}  \\
\midrule
{\model} 
& {49.6} & {\textbf{53.9}} 
& {\textbf{74.8}} & {\textbf{78.7}} 
& {\textbf{42.2}} & {\textbf{51.9}} 
& {58.8} & {\textbf{61.5}}  \\
\bottomrule
\end{tabular}}

\end{table*}

\begin{table*}[t]\centering
\setlength{\tabcolsep}{3.5mm}{

\caption{The Impact of LLMs on Utility Modeling.
} 
\label{tab:llm_impact_on_utility}
\begin{tabular}{c|c|c|c|c|c|c|c|c}

\toprule
\multirow{2}{*}{LLMs}
& \multicolumn{2}{c|}{NQ} 
& \multicolumn{2}{c|}{TriviaQA} 
& \multicolumn{2}{c|}{HotpotQA} 
& \multicolumn{2}{c}{PopQA} 
\\    
\cmidrule{2-9}
 {} 
 & {EM} & {F1} 
 & {EM} & {F1} 
 & {EM} & {F1} 
 & {EM} & {F1}\\ 
\midrule 
{Baichuan2$_{\textsc{7B}}$} 
& {32.1} & {36.8} 
& {63.5} & {67.8}  
& {25.3} & {33.2}   
& {23.6} & {25.1}   \\
{Llama2$_{\textsc{7B}}$} 
& {42.1} & {46.5} 
& {69.1} & {73.5} 
& {35.7} & {45.2}  
& {\textbf{55.7}} & {\textbf{56.8}}   \\
{Llama2$_{\textsc{13B}}$} 
& {\textbf{45.2}} & {\textbf{49.8}} 
& {\textbf{71.9}} & {\textbf{76.0}} 
& {\textbf{38.3}} & {\textbf{47.9}}  
& {55.1} & {56.4}  \\
\bottomrule
\end{tabular}}

\end{table*}

\subsubsection{Main Results}
From the empirical results across multiple knowledge-intensive datasets, the major findings can be summarized as follows:
\begin{itemize}[leftmargin=*]
\item \textbf{Retrieval largely improves performance.}
When compared against approaches without retrieval, the retrieval-augmented approaches demonstrated the superiority in EM metric even when compared to the strong baseline ChatGPT, showing the tremendous power and potential of retrieval-augmented generation. Besides, with retrieval-augmentation, even small-scale LLMs (\eg 7B LLMs) can achieve comparable performance \wrt large-scale LMs (\eg 13B LLMs) in terms of EM metric, showing their power in pursuit of true knowledge.
\item \textbf{Long-tail queries benefit more from retrieval.}
As depicted in Table \ref{tab:overall_result1}, retrieval-augmented approaches achieve most performance gain in PopQA dataset with long-tail queries while the large-scale competitive LLM ChatGPT performs worst in this dataset due to the untimely updated knowledge, which illustrates that performance degradation due to knowledge updating issues can be alleviated to a great extent by retrieval.
\item \textbf{Supervised Fine-tuning improves instruction following.}
With regard to F1 metric, we find that approaches trained with retrieved passages perform better, showcasing their ability in instruction following for abstracting concise answers (which is the case of ChatGPT with remarkable instruction following ability for zero-shot tasks). However, for approaches without supervised fine-tuning, there exists a \textit{seesaw effect} between EM and F1 with one focuses on answer accuracy while other focuses on the balance between exactness and conciseness (which is stated in input prompts).
\item \textbf{The way of incorporating external information matters!}
Different from approaches that directly incorporate the retrieved passages for answer generation, our proposed {\model} endows multi-layered thoughts to \textit{``take the essence and discard the dross''} so that the most useful information of retrieved passages can be abstracted and distraction information can be dropped for end tasks. Experimental results in four different datasets illustrate and verify the rationale of our proposed {\model}.
\end{itemize}

\subsubsection{Analysis}\label{sec:exp_analysis}

\begin{figure}[t]
	\centering
	\subfigure[EM]{\includegraphics[width=0.48\columnwidth]{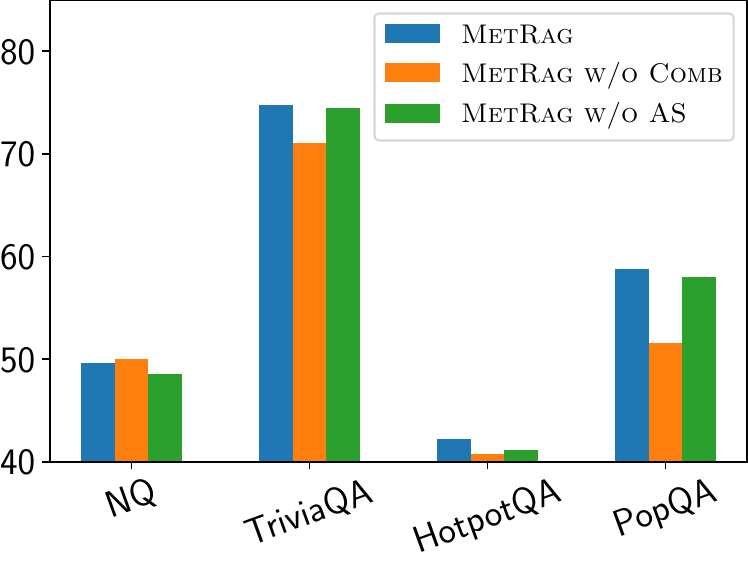}}
	\subfigure[F1]{\includegraphics[width=0.48\columnwidth]{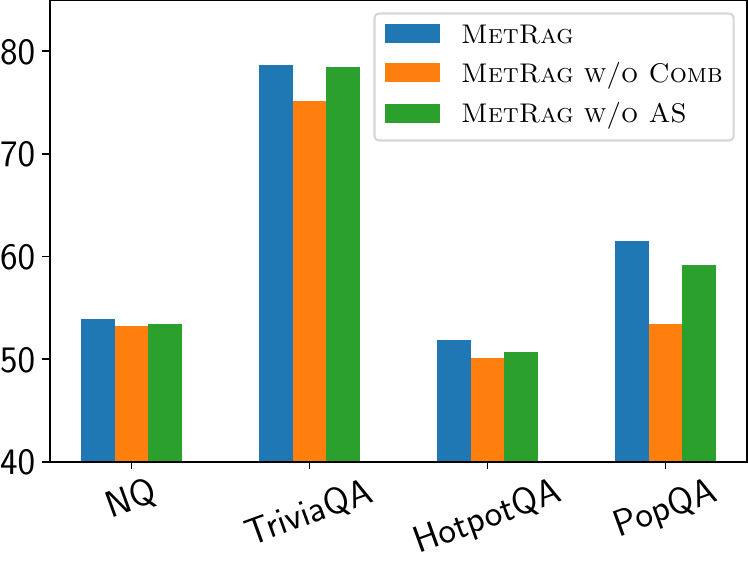}}
  \caption{Ablation study.}
	\label{fig:ab_study}
\end{figure}

\paragraph{Ablation study.} We examine the effectiveness of each component in {\model} by preparing the following variants: 
i) {\comb}, which removes the combination of similarity model and utility model and degrades to the original similarity based retriever;
ii) {\as}, which removes the task-adaptive summarization for information integration.
We plot the performance comparison in Figure \ref{fig:ab_study}, from which we can observe that the overall performance would drop a lot when either component is discarded, thus the effectiveness of our dedicate design is verified. 
In particular, we find that the {\comb} performs worst among all variants, showcasing that the augmented information, which serves as the cornerstone in retrieval-augmented generation, deserves more attention.

\begin{figure}[t]
    \centering
    \includegraphics[width=0.9\columnwidth]{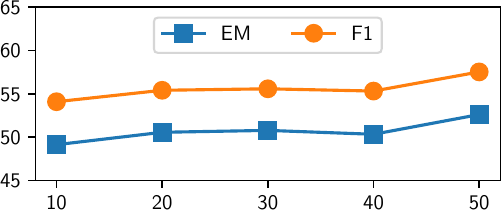}
    \caption{The influence of passage window size.}
    \label{fig:topk}
\end{figure}

\paragraph{Passage window size matters in utility model performance!}
Since the utility model plays a key role in bring in the supervision of LLMs into passage selection process, we take a closer look at how the passage window size $\left\| \mathcal{D}^\bullet \right\|$ \footnote{Here, since the empty string $es$ is added by default, we only count the passages involved in training.} would influence the final performance by directly incorporating the utility model trained under different settings to the final task. Due to the space limit, we average the metrics in four datasets and present the result in Figure \ref{fig:topk} and find that the model performance improves as the window size grows, which demonstrates that the growing passage window size endows more LLMs' powers for distinguishing passage significance among diverse inputs to the utility model thus improving performance on downstream tasks. However, due to the computational burden (the model training time grows linearly with the passage window size) that a large passage window size imposes during training, there is need to balance the trade-off between performance and cost.

\paragraph{Impact of selective retrieval.}\label{sec:exp_analysis_selective_retrieval}
When optimizing the utility model, we add an empty string $es$ in the training process to leverage the knowledge of LLMs to achieve selective retrieval. Here, we conduct some case studies to examine this mechanism. 
As we can see from Table \ref{tab:selective_retrieval}, there are example queries that the utility model deems no retrieval and in total there are 20.9\% queries which the utility model ranks the empty string $es$ higher than other documents. As we can see from Table \ref{tab:selective_retrieval}, many of the listed queries are commonsense knowledge that has been memorized in LLMs' parameters and we can easily call an LLM for the required answer instead of retrieval, which illustrates that in retrieval-augmented generation, despite the further knowledge external corpus introduces, the inherent knowledge of LLMs when deploying retrieval-based techniques is worth investigation and we believe that our work has provided a straightforward and plausible solution.
However, this design still has certain limitations, \ie the LLM used for end task needs to be the same or stronger than the one utilized during utility model training, otherwise this mechanism may not work.

\begin{myverbbox}[\small]{\VerbContentQueriesD}
Is Cartagena or Chess more popular around the world? 
In what country is Chalhuacocha? 
What was Ian Fleming’s first Bond book?
When was catch me if you can made?
Cathay is a poetic name for which country?
Who sings sugar sugar you are my candy girl?
Who used to present I'm a celebrity now?
What county is Icknield Walk First School located in?
Where does the cell spend most of its time in the cell cycle?
What country lies to north of the Republic of Chad?
Are Vintage Life and InStyle both US-based magazines?
Who is the first president to be impeached?
Are Andrew Stevens and Charles Burnett both American?
Which Genre of Television did Charles Quinton Murphy act for?
How was President Kennedy assassinated?
Which animal is the carrier of the h1n1 virus?
What part of the body produces insulin?
\end{myverbbox}

\begin{table}[H]
	\centering
	\caption{Queries Deemed No Retrieval by Utility Model.}
	\label{tab:selective_retrieval}
	\resizebox{\linewidth}{!}{\begin{tabular}{c}
	\toprule
        \VerbContentQueriesD       \\ \midrule
	\end{tabular}}
\end{table}

\paragraph{The impact of LLMs on utility modeling.}
To further analyse the impact of LLMs on utility modeling, beside Llama2$_{\textsc{13b}}$ we also add Baichuan$_{\textsc{7b}}$ and Llama2$_{\textsc{7b}}$ as the base LLMs for comparison by directly incorporating the utility model trained under different settings to the final task and present the metrics in four datasets in Table \ref{tab:llm_impact_on_utility}, from which we have two main conclusions: 1) Usually, a large-scale LLM can outperforms an LLM of smaller-scale, \eg Llama2$_{\textsc{13b}}$ vs Llama2$_{\textsc{7b}}$. Furthermore, we also find an interesting phenomenon that sometimes (\eg PopQA dataset) the score of Llama2$_{\textsc{7b}}$ is higher than that of Llama2$_{\textsc{13b}}$, which is consistent of the findings reported in Self-RAG\cite{selfrag2023asai}. 2) Baichuan$_{\textsc{7b}}$ underperforms the Llama series in our experiments, one possible reason is that it places greater emphasis on its application in specific languages such as Chinese, as is stated in its original paper\cite{baichuan2023baichuan2}.

\paragraph{Investigation into task-adaptive summarizer.}
To start with, we introduce a case study to illustrate the virtues of our task-adaptive summarizer in Table \ref{tab:summary_case}, as we can see from the table, since the original information is quite long and with a lot of distracting information, an LLM tend to get lost in a large chunk of words. However, with our task-adaptive summarizer, the most relevant information for this query is extracted thus an LLM can easily answer the question given the extracted knowledge.

\begin{myverbbox}[\small]{\VerbContentQueD}
What is Walter de la Pole's occupation?
\end{myverbbox}

\begin{myverbbox}[\small]{\VerbContentOriD}
Walter de la Pole: Family  Walter was the son and heir of the MP 
Sir Edmund de la Pole and his second wife. 
Walter de la Pole  Sir Walter de la Pole (November 1371 – 1434), of 
Dernford in Sawston, Cambridgeshire, was an English politician.
Richard de la Pole  Richard de la Pole (1480 – 24 February 1525)
was a pretender to the English crown. Commonly nicknamed "White 
Rose", he was the last Yorkist claimant to actively and openly seek
the crown of England. He lived in exile...(4203 characters)
\end{myverbbox}

\begin{myverbbox}[\small]{\VerbContentSumD}
Walter de la Pole was an English politician. (44 characters)
\end{myverbbox}

\begin{table}[h]
	\centering
	\caption{Case Study: Task-adaptive Summarizer Output.}
	\label{tab:summary_case}
	\resizebox{\linewidth}{!}{\begin{tabular}{c}
	\toprule
        Query \\ \midrule
        \VerbContentQueD       \\ \midrule
        Original information \\ \midrule
        \VerbContentOriD       \\ \midrule
        Task-adaptive summary \\ \midrule
        \VerbContentSumD       \\ \bottomrule
	\end{tabular}}
\end{table}

In sum, the virtues of task-adaptive summarizer are two-fold. 
On the one hand, by using a small-scale LLM (\ie a 13B Llama model) as the summarizer it could speed up the inference for downstream tasks. 
The summary ratio across the four datasets are 8.25\%, 7.35\%, 6.77\% and 7.32\%
respectively, which might increase a small amount of inference cost by adding the summary stage, nevertheness, since the inference cost of LLMs is linearly correlated with token number, with the knowledge derived from the summarizer, it can potentially decrease the inference cost for high-cost commercial LLMs such as GPT4. 
On the other hand, it further enhances the performance by extracting the most relevant information for a given task so as to alleviate distracting information, which improves the EM and F1 metrics across the four datasets on average by 1.4\% and 1.7\% respectively, which further justifies our model design.

\section{Conclusion}\label{sec:con}
In this work, we propose {\model}, a multi–layered thoughts enhanced retrieval-augmented generation framework, which first embraces LLM's supervision to obtain utility-oriented thoughts and combines the similarity and utility of documents for performance boosting, and further pursuits compactness-oriented thoughts via a task-adaptive summarizer.
Finally, with multi-layered thoughts from the precedent stages, an LLM is called for knowledge-augmented generation.
Extensive experiments on knowledge-intensive tasks have demonstrated the superiority of the proposed {\model}.

\section{Limitations}\label{sec:limits}
One of the limitation of our work is that the effectiveness of utility model is highly hinges on a strong LLM's supervision, although we have find that an LLM like LLama 7B or 13B is enough for training a satisfactory utility model. In sum, our work opens up a fresh perspective to reconsider retrieval-augmented generation, but more complex situation that require reading a large amount of material to
answer (\eg legal or medical documents) is still unresolved. Hence, extending our framework in the super-long contexts is one of the future work.

\clearpage
\bibliography{anthology,custom}
\bibliographystyle{acl_natbib}

\clearpage
\appendix

\section{Appendix}
\label{sec:appendix}
\subsection{Prompt Template}\label{app:prompt}
\subsubsection{Utility Model Training Prompt}
When training the utility model, we design different prompts so that an LLM can output its perplexity when i) answering the question with retrieved document; ii) answering the question directly. We present the prompts in Table \ref{tab:utility_prompt}.

\begin{myverbbox}[\small]{\VerbContentWithInfoD}
Please answer the question based on the given context. Question: 
[question] The context related to the question is as follows:
[retrieved document]. Answer: [answer] 
\end{myverbbox}
\begin{myverbbox}[\small]{\VerbContentNoInfoD}
Please answer the question. Question: [question] Answer: [answer]
\end{myverbbox}

\begin{table}[H]
	\centering
	\caption{Utility Model Training Prompt.}
	\label{tab:utility_prompt}
	\resizebox{\linewidth}{!}{\begin{tabular}{c}
	\toprule
        w/ retrieved document \\ \midrule
        \VerbContentWithInfoD       \\ \midrule
        w/o retrieved document \\ \midrule
        \VerbContentNoInfoD       \\ \midrule
	\end{tabular}}
\end{table}

\subsubsection{Task-adaptive Summarizer Training and Inference Prompt}
To endow retrieval-augmented generation with compactness-oriented thought, we carefully train a task-adaptive summarizer for information integration and extraction, where we show the prompt in Table \ref{tab:summary_prompt}.

\begin{myverbbox}[\small]{\VerbContentSummaryD}
Instruction:
You are an excellent summary generation robot. Given the following 
question (Question) and texts (Docs), you need to summarize these 
texts (Docs) into a concise abstract to adequately address the 
corresponding question.
Question:
[question], [options if it is multiple-choice question]
Docs:
[retrieved documents]
Summary:
\end{myverbbox}

\begin{table}[H]
	\centering
	\caption{Task-adaptive Summarizer Prompt.}
	\label{tab:summary_prompt}
	\resizebox{\linewidth}{!}{\begin{tabular}{c}
	\toprule
        \VerbContentSummaryD       \\ \midrule
	\end{tabular}}
\end{table}

\subsubsection{Knowledge-augmented Generation Prompt}
We show the prompt for generating the final answer in Table \ref{tab:qa_task}.

\begin{myverbbox}[\small]{\VerbContentQAD}
Instruction:
You are an AI assistant for answering questions. Based on the given 
question (Question) and the corresponding information (Info), please 
provide the correct answer as concise as possible according to the 
info and your commonsense.
Info:
[retrieved documents / summary]
Question:
[question]
Answer:
Please answer the question in the form of 2 or 3 words.
\end{myverbbox}

\begin{table}[H]
	\centering
	\caption{Knowledge-augmented Generation Prompt.}
	\label{tab:qa_task}
	\resizebox{\linewidth}{!}{\begin{tabular}{c}
	\toprule
        \VerbContentQAD       \\ \midrule
	\end{tabular}}
\end{table}

\clearpage
\end{document}